\documentclass{article}
\usepackage{ijcai25}

\usepackage{times}
\usepackage{soul}
\usepackage{url}
\usepackage[hidelinks]{hyperref}
\usepackage[utf8]{inputenc}
\usepackage[small]{caption}
\usepackage{graphicx}
\usepackage{amsmath}
\usepackage{amsthm}
\usepackage{booktabs}
\usepackage{algorithm}
\usepackage{algpseudocode}
\usepackage[switch]{lineno}
\usepackage{graphicx}
\usepackage{changepage} 
\usepackage{multirow}
\usepackage{xcolor}         
\usepackage{float}
\usepackage{ragged2e}
\usepackage{subcaption}
\usepackage{newfloat}
\usepackage{listings}
\usepackage{color}
\usepackage{placeins}
\usepackage{pifont}
\usepackage{array}
\definecolor{jsonkeycolor}{rgb}{0,0.5,0}
\definecolor{jsonstringcolor}{rgb}{0.6,0.6,1}
\definecolor{jsonnumbercolor}{rgb}{0.8,0.5,0.5}

\lstdefinelanguage{json}{
    basicstyle=\small\ttfamily,
    showstringspaces=false,
    breaklines=true,
    frame=none,
    backgroundcolor=\color{white},
    numbers=none, 
    literate=
     *{0}{{{\color{jsonnumbercolor}0}}}{1}
      {1}{{{\color{jsonnumbercolor}1}}}{1}
      {2}{{{\color{jsonnumbercolor}2}}}{1}
      {3}{{{\color{jsonnumbercolor}3}}}{1}
      {4}{{{\color{jsonnumbercolor}4}}}{1}
      {5}{{{\color{jsonnumbercolor}5}}}{1}
      {6}{{{\color{jsonnumbercolor}6}}}{1}
      {7}{{{\color{jsonnumbercolor}7}}}{1}
      {8}{{{\color{jsonnumbercolor}8}}}{1}
      {9}{{{\color{jsonnumbercolor}9}}}{1}
      {:}{{{\color{jsonkeycolor}{:}}}}{1}
      {,}{{{\color{jsonkeycolor}{,}}}}{1}
      {\{}{{{\color{jsonkeycolor}{\{}}}}{1}
      {\}}{{{\color{jsonkeycolor}{\}}}}}{1}
      {[}{{{\color{jsonkeycolor}{[}}}}{1}
      {]}{{{\color{jsonkeycolor}{]}}}}{1},
}
\floatstyle{ruled}
\newfloat{listing}{tb}{lst}{}
\floatname{listing}{Listing}

\title{Automating Intervention Discovery from Scientific Literature: A Progressive Ontology Prompting and Dual-LLM Framework\thanks{Corresponding author: Jinjun Xiong (jinjun@buffalo.edu)}}
\author{
Yuting Hu$^1$
\and
Dancheng Liu$^1$\and
Qingyun Wang$^2$\and
Charles Yu$^2$\and
Chenhui Xu$^1$\and
Qingxiao Zheng$^1$\and
Heng Ji$^2$\And
Jinjun Xiong$^{1}$\\
\affiliations
$^1$University at Buffalo\\
$^2$University of Illinois at Urbana-Champaign\\
\emails
\{yhu54,dliu37,cxu26,qingxiao,jinjun\}@buffalo.edu,
\{qingyun4,ctyu2,hengji\}@illinois.edu
}

\begin{document}

\maketitle

\begin{abstract}
    Identifying effective interventions from the scientific literature is challenging due to the high volume of publications, specialized terminology, and inconsistent reporting formats, making manual curation laborious and prone to oversight. To address this challenge, this paper proposes a novel framework leveraging large language models (LLMs), which integrates a progressive ontology prompting (POP) algorithm with a dual-agent system, named LLM-Duo. On the one hand, the POP algorithm conducts a prioritized breadth-first search (BFS) across a predefined ontology, generating structured prompt templates and action sequences to guide the automatic annotation process. On the other hand, the LLM-Duo system features two specialized LLM agents, an explorer and an evaluator, working collaboratively and adversarially to continuously refine annotation quality. We showcase the real-world applicability of our framework through a case study focused on speech-language intervention discovery. Experimental results show that our approach surpasses advanced baselines, achieving more accurate and comprehensive annotations through a fully automated process. Our approach successfully identified 2,421 interventions from a corpus of 64,177 research articles in the speech-language pathology domain, culminating in the creation of a publicly accessible intervention knowledge base \footnote{Project website: \url{https://slp.xlabub.com/}} with great potential to benefit the speech-language pathology community.
\end{abstract}
\section{Introduction}
Evidence-based interventions refer to practices and treatments grounded in systematic research and proven effective through controlled studies \cite{rutten2021evidence}\cite{melnyk2022evidence}. It emphasizes the use of evidence from well-designed and well-conducted research as the foundation for healthcare decision-making \cite{sackett1997evidence}. Intervention discovery from scientific literature enables researchers to keep abreast of the latest advancements and facilitate valuable insights that can significantly enhance the healthcare quality \cite{usai2018knowledge}\cite{wang2023scientific}. However, due to the labor-intensive nature of human review, only a small fraction of intervention knowledge is systematically collected and curated. In healthcare, one of the biggest challenges for healthcare providers is the efficient identification of relevant intervention evidence from an overwhelming body of research, highlighting the urgent need for automated knowledge extraction tools to streamline the process and enhance the accessibility of this valuable information.

In recent years, large language models (LLMs) have been employed to categorize research papers, extract key findings, summarize complex studies, and create conversational assistants for question-answering and note generation, showing their impressive ability in understanding and extracting valuable insights from text \cite{achiam2023gpt}\cite{li2024preliminary}. Many studies have utilized LLMs to streamline various subtasks involved in knowledge graph construction, such as named entity recognition (NER), relation extraction (RE), event extraction (EE), and entity linking (EL) \cite{wang2023gpt}\cite{zhu2024llms}. Some research has also explored the collaboration between LLMs and human annotators to improve annotation quality \cite{kim2024meganno+}\cite{wang2024human}\cite{tang2024pdfchatannotator}. However, extracting intervention knowledge from long-range, domain-specific literature remains a significant challenge. On the one hand, developing human-annotated datasets for training deep learning models in NER and RE tasks requires specialized domain expertise to accurately interpret the literature. On the other hand, mining knowledge from long-range documents is a great challenge due to the vast volume of content, the inherent ambiguity of natural language, and the individual bias of human interpretation \cite{ye2022generative}. Particularly in healthcare contexts, these challenges are further compounded by the need for specialized therapeutic expertise, labor-intensive manual annotation, and difficulties in maintaining consistency and scalability \cite{zhao2021novel}. In this context, LLMs offer a promising alternative through in-context learning, enabling scalable information extraction without the need for extensive human-labeled data. Despite these advancements, fully automated knowledge graph construction remains a challenge, particularly when dealing with long-range documents. Most current knowledge graph construction approaches focus on short texts, leaving significant potential for further development in handling more complex, lengthy content.

In this paper, we address the challenge of automating intervention discovery via LLMs by formulating it as a prompt design and annotation scheduling problem with a predefined intervention ontology graph structure and designing a framework leveraging two LLM agents to iteratively enhance the annotation quality. Specifically, we introduce a progressive ontology prompting (POP) algorithm that employs an outdegree-prioritized breadth-first search (BFS) across the intervention ontology to create a series of prompt templates and action sequences to guide the annotation process conducted by LLMs. To enhance the annotation quality, we propose LLM-Duo, an interactive annotation framework by leveraging the power of LLMs while addressing the limitations of LLMs. Particularly, it integrates two LLM agents working both collaboratively and adversarially to refine annotation generation.

To showcase the practical impact of our approach, we apply our method in a case study of speech-language intervention discovery. We conduct experiments to compare our intervention discovery framework with several advanced baselines including long context LLM (i.e., GPT-4-Turbo with 128k context window length), and RAG-based annotation chatbot with advanced prompting techniques including Chain-of-Thought (CoT) \cite{wei2022chain} and Self-Refine \cite{madaan2024self}. The experimental results demonstrate that our method not only delivers more accurate and comprehensive annotations over these strong baselines but also significantly accelerates the intervention discovery process. Furthermore, through our framework, we successfully curate a speech-language intervention knowledge base, providing a valuable resource for the speech-language pathology community. To our knowledge, this is the first intervention knowledge base in the speech-language pathology field.

\noindent\textbf{Related Work.} Traditional approaches to automated knowledge discovery typically rely on pipelines to handle various NLP tasks such as named entity recognition, relation extraction, coreference resolution, entity linking, and event detection \cite{luan2018multi}\cite{martins2019joint}\cite{wei2019novel}\cite{zhong2023comprehensive}\cite{laurenzi2024llm}. Recent advancements leverage LLMs to generate relational triplets in zero/few-shot settings for knowledge graph construction, achieving promising results \cite{wei2023zero}\cite{sun2024consistency}\cite{he2024zero}. Some studies \cite{zhang2024extract}\cite{carta2023iterative}\cite{vamsi2024human}\cite{zhu2024llms} have further streamlined knowledge graph construction by breaking it down into distinct phases, enabling LLMs to infer knowledge graph schemas without relying on predefined ontologies. However, these methods are often constrained to short texts or have only been validated on tasks like entity and relation extraction using human-annotated datasets, such as DuIE2.0 \cite{li2019duie} and DocRED \cite{yao2019docred}, without being proven effective in real-world applications. Moreover, domain-specific knowledge often exhibits complex patterns that cannot be captured solely through sentence-level syntactic structures. As a result, most existing approaches \cite{du2020knowledge}\cite{rossanez2020kgen}\cite{alam2023automated} are limited to handling abstracts and fail to extract and summarize knowledge across long-range contexts.
\section{Preliminaries}
A knowledge graph (KG) is a semantic network structured as an ontology, consisting of concepts and their relationships in a clear, interpretable format at scale \cite{peng2023knowledge}. For intervention knowledge discovery from literature, LLMs can enhance this process by leveraging their capabilities to understand long-range text. This allows for transforming unstructured data into structured formats, and finally, populating the intervention ontology to create the intervention knowledge graph.

In our methodology, the intervention KG ontology is crafted by domain experts, which can be represented by a directed acyclic graph (DAG) $\mathcal{G}=(\mathcal{E},\mathcal{R}, \mathcal{F})$. Here $\mathcal{E}$, $\mathcal{R}$, and $\mathcal{F}$ are sets of concepts, relationships, and semantic triples respectively. $\mathcal{F}$ is a collection of triples $(h, r, t)$ with a head concept $h\in \mathcal{E}$, a tail concept $t\in \mathcal{E}$, and a relation $r\in \mathcal{R}$ \cite{gruninger1995methodology}. To effectively instruct LLMs to extract intervention knowledge anchored to $\mathcal{G}$ automatically, the design of annotation prompts and the query sequences plays a crucial role. We thereby frame the problem of automated intervention knowledge discovery via LLMs as one of prompt design and scheduling, described by the following equation:
\begin{equation}
\small f(\mathcal{G}(\mathcal{E},\mathcal{R}, \mathcal{F}))=\left \{(Prompt_i, Order_i)|i\in [1, N] \right \}
\end{equation}
where $f$ is a function that translates intervention KG ontology into a set of annotation prompts and query sequences for the LLMs. A common case of $f$ is directly prompting LLMs to generate triplets in a zero-shot/few-shot manner by including the whole KG schema within the prompt such as the annotation methods used in \cite{mihindukulasooriya2023text2kgbench}\cite{kommineni2024human}. However, those methods generate annotations in one shot and ignore the importance of contextual correlations between concepts within their surrounding neighborhood, resulting in incomplete annotations.

\begin{figure*}
  \centering
  \includegraphics[width=1.0\textwidth]{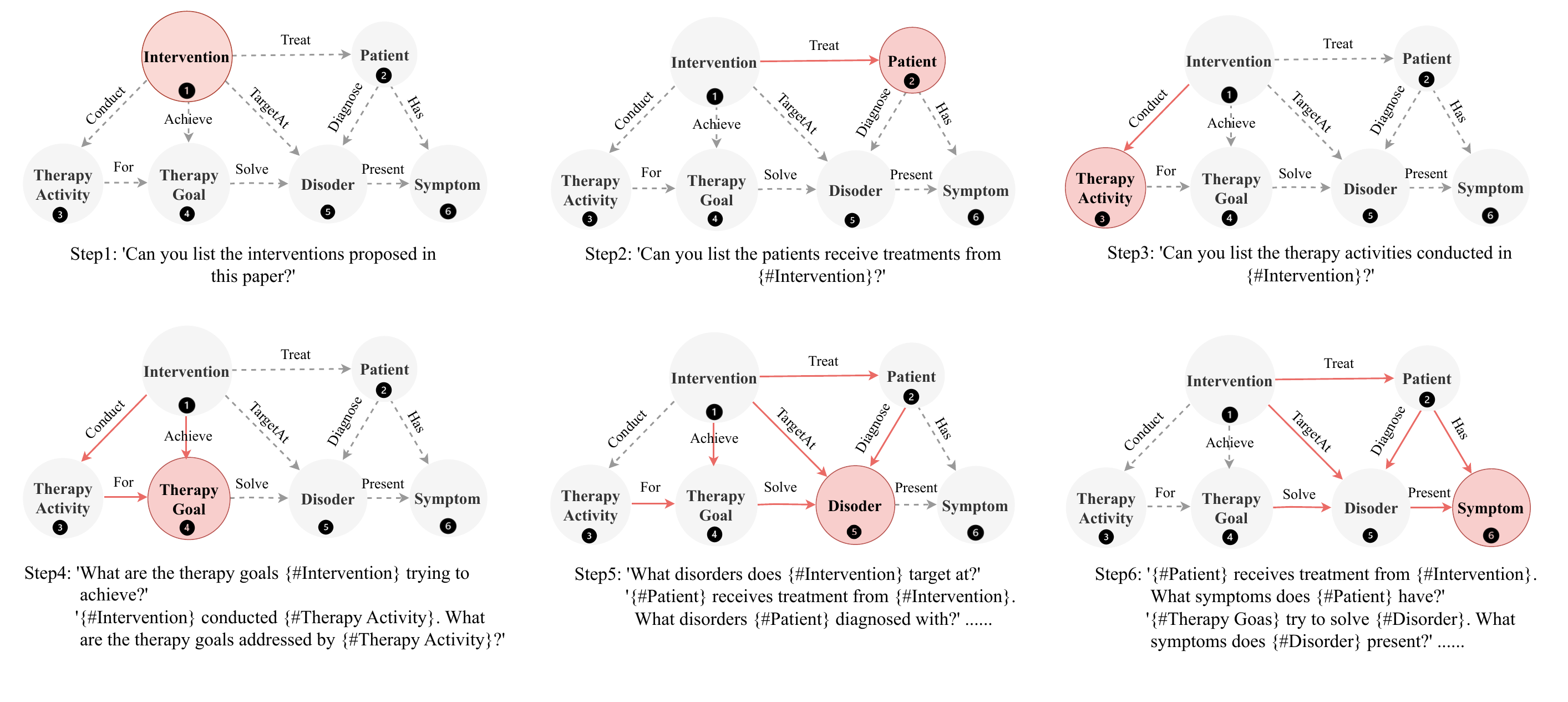}
  \caption{Illustration of prompt design and scheduling based on the progressive ontology prompting algorithm.}
  \vspace{-3mm}
  \label{fig:bfsprompt}
\end{figure*}

\section{Methodology}
In this section, we first introduce the POP algorithm that converts an intervention KG ontology into a set of annotation prompt templates and query orders, then propose an interactive annotation framework based on two LLM agents to enable more convincing and accurate annotation generations.

\subsection{Progressive Ontology Prompting}
We develop a progressive ontology prompting (POP) algorithm that employs a prioritized BFS on the intervention ontology graph $\mathcal{G}(\mathcal{E},\mathcal{R}, \mathcal{F})$ to generate a set of annotation prompt templates and query sequences for LLMs. In our algorithm, the prompt formulation and scheduling follow a progressive manner. As illustrated in Figure~\ref{fig:bfsprompt}, the annotation process begins at a source node (i.e., a concept node that only has outgoing edges) and continues by traversing its neighboring nodes in the order of a prioritized BFS. To allow for quick accessing a large portion of the graph, we enhance BFS by sorting neighboring nodes based on their out-to-in ratio $R(v)$, which is defined by:
\begin{equation}
\small R(v) = \frac{|\left\{ (h,r,t)\in \mathcal{F} \right\}|h=v|}{|\left\{ (h,r,t)\in \mathcal{F} \right\}|t=v|}
\end{equation}

Our algorithm selects the neighboring node with the maximum $R$ value to visit in the next step. For instance, in the example of Figure~\ref{fig:bfsprompt}, visiting the `Patient' node before the `Disorder' node can provide more context for the `Disorder' concept annotation. For each concept node $v$, we use the visited nodes within its $k$ hop neighborhood as its context. The $Prompt_v$ for annotating concept $v$ is crafted based on its context and the completed annotations within that context. The action order $Order_v$ for $Prompt_v$ is determined by the sequence in which node $v$ is visited during the prioritized BFS traversal. 

Our algorithm first follows prioritized BFS traversal to capture the local $k$ hop context and visit order of a specific concept node $v$, then composes the annotation prompts $Prompt_v$ based on its ontology substructure $N_{k}(v)$ and completed annotations within its context. This process can be expressed as follows:
{\scriptsize
\begin{align}
    Prompt_v &\gets T_v(Annotation(N_{k}(v))) \\
    T_v &\gets \left\{ Prefix\left( N_{k-1}(u) \right) \oplus \right. \nonumber \\
        &\hspace{1em} \left. Question\left( (v,e,u) \mid (v,e,u) \in F \right) \mid u \in N_1(v) \right\}
\end{align}}
, where $\oplus$ is the concatenation. $Prompt_v$ represents a set of annotation prompts for node $v$, generated by applying completed annotations to the prompt template $T_v$. As illustrated in Figure~\ref{fig:bfsprompt}, the prompt template $T_v$ consists of two parts: 1) \textit{Question}, an annotation question derived from the relationship between node $v$ and one of its neighboring nodes $u$; and 2) \textit{Prefix}, a description based on the $k-1$ hop path of neighbor node $u$. We leverage few-shot learning to task LLMs in generating prompt templates.
\subsection{LLM-Duo Annotation Framework}
\begin{figure}[h]
\centering
  \includegraphics[width=6cm]{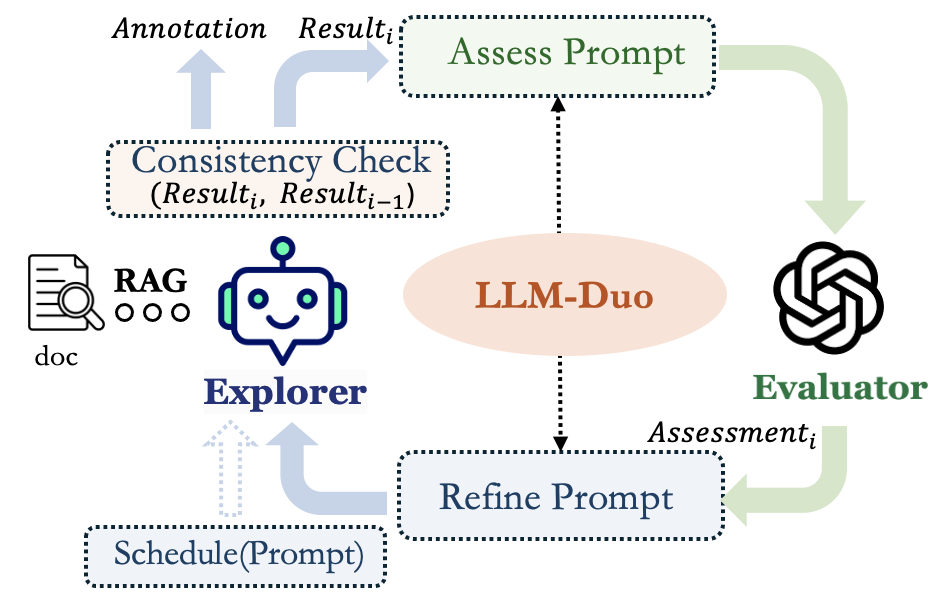}
  \caption{Iterative annotation with two LLM agents under the LLM-Duo framework.}
  \vspace{-5mm}
  \label{fig:llmduo}
\end{figure}
\begin{figure*}
    \centering
    \includegraphics[width=1.0\textwidth]{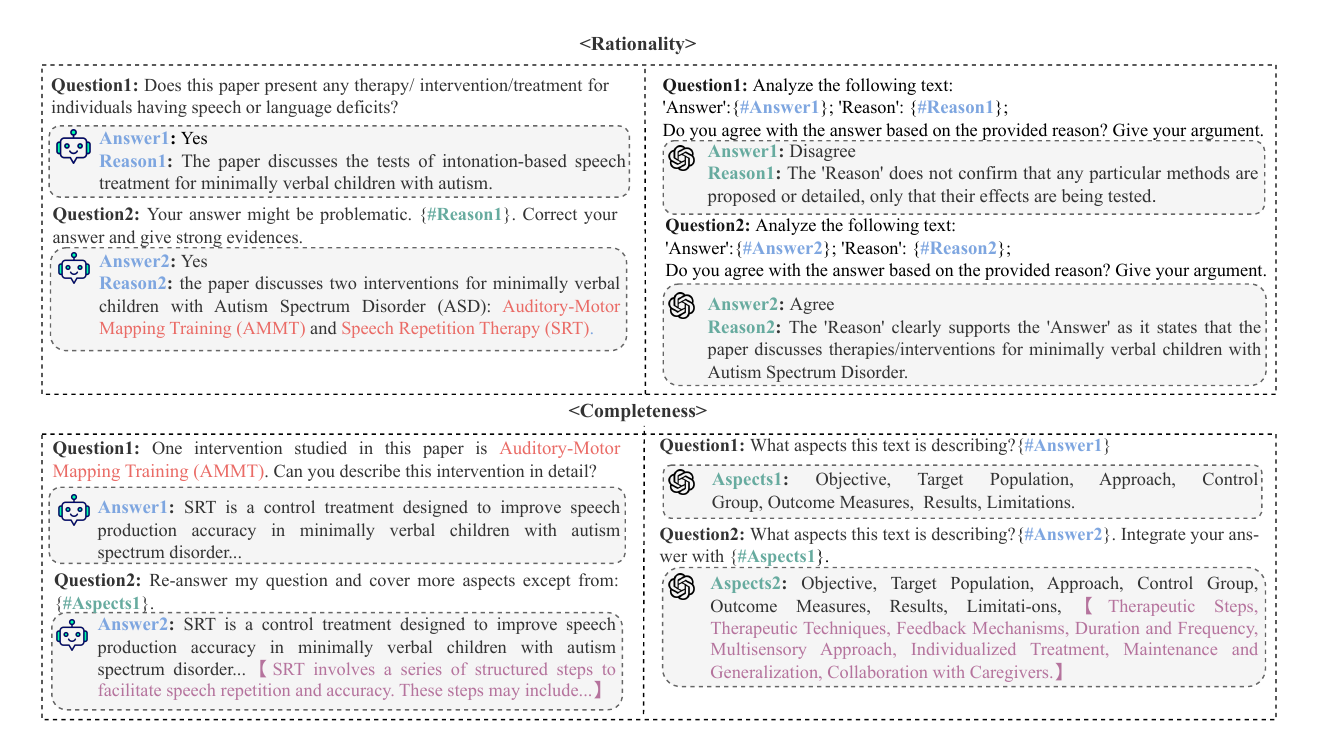}
    \caption{Annotation examples of speech-language intervention discovery using the LLM-Duo framework.}
    \label{fig:example}
    \vspace{-5mm}
\end{figure*}
To guarantee the integrity and reliability of LLM annotations, we propose LLM-Duo, a dual-agent annotation system. The central idea of a multi-agent system is to employ combinations of LLMs that can converse with each other to collaboratively accomplish tasks \cite{wu2023autogen}. Drawing inspiration from the multi-agent debate idea in \cite{kim2024can}, we developed a framework where agents work both collaboratively and adversarially to enhance the quality of annotations. 

The architecture of LLM-Duo is shown in Figure~\ref{fig:llmduo}, featuring two LLM agents: the \textit{explorer} and the \textit{evaluator}. Specifically, the \textit{explorer} is a chatbot performing annotation tasks using zero-shot question answering (QA). To break the context window limit of LLMs and ensure the generated annotations are faithful to the provided literature content, RAG is employed in \textit{explorer} to reference relevant sources, minimizing LLM hallucinations. To further improve the accuracy and reliability of the annotations, the \textit{evaluator} LLM is incorporated to review and validate the \textit{explorer's} responses, ensuring higher-quality results.

LLM-Duo will be tasked with annotation prompts following the sequential order generated by the POP algorithm. During each annotation cycle for a specific concept node, when focusing on concepts that emphasize rationality (e.g., disorder, intervention efficacy), the \textit{explorer} provides an answer and an explanatory rationale to the \textit{evaluator}. The \textit{evaluator} then reviews the reasoning and offers feedback. Based on this feedback, the \textit{explorer} either refines its answer or, if in disagreement, presents stronger evidence to defend its original answer and challenge the \textit{evaluator's} judgment. For concepts that emphasize completeness (e.g., intervention procedure, therapy activity), the \textit{evaluator} extracts the aspects covered in each round of the \textit{explorer's} answer, combines them with aspects from previous rounds, and prompts the \textit{explorer} to expand further beyond the newly integrated aspect collection. This iterative process continues until the annotations reach a consistent and comprehensive state. As the example shown in Figure~\ref{fig:example}, by facilitating interactive loops between two LLM agents, LLM-Duo enables more accurate and complete annotations. 
\section{Experiments}
\subsection{Implementation}
For LLM-Duo, the \textit{explorer} is a chatbot built on LLM with RAG, implemented with Llamaindex\footnote{\scriptsize{\url{https://www.llamaindex.ai}}} framework. We use OpenAI `text-embedding-3-large'\footnote{\scriptsize{\parbox{\dimexpr\linewidth-1em}{\url{https://platform.openai.com/docs/guides/embeddings/embedding-models}}}} as the embedding model and set the chunk size to 256 tokens with an overlapping size of 128. Particularly, we use `FastCoref' \cite{otmazgin2022f} to process text chunks for coreference resolution before text embedding. Additionally, we include the document ID as metadata for chunks and apply a metadata filter in the chat engine to ensure that the \textit{explorer} only answers based on the specific document being annotated. We use Chroma\footnote{\scriptsize{\url{https://github.com/chroma-core/chroma}}} as the vector database. We set the retrieval to be on the top 8 text chunks based on similarity scores reranked with SentenceTransformerRerank\footnote{\scriptsize{\parbox{\dimexpr\linewidth-1em}{\url{https://docs.llamaindex.ai/en/stable/examples/node_postprocessor/SentenceTransformerRerank}}}} employing the `cross-encoder/ms-marco-MiniLM-L-2-v2'\footnote{\scriptsize{\parbox{\dimexpr\linewidth-1em}{\url{https://huggingface.co/cross-encoder/ms-marco-MiniLM-L-2-v2}}}} model in Llamaindex. The evaluator is an external LLM who does not share any document context with the \textit{explorer}.
\begin{figure}[ht]
    \centering
  \includegraphics[width=7cm]{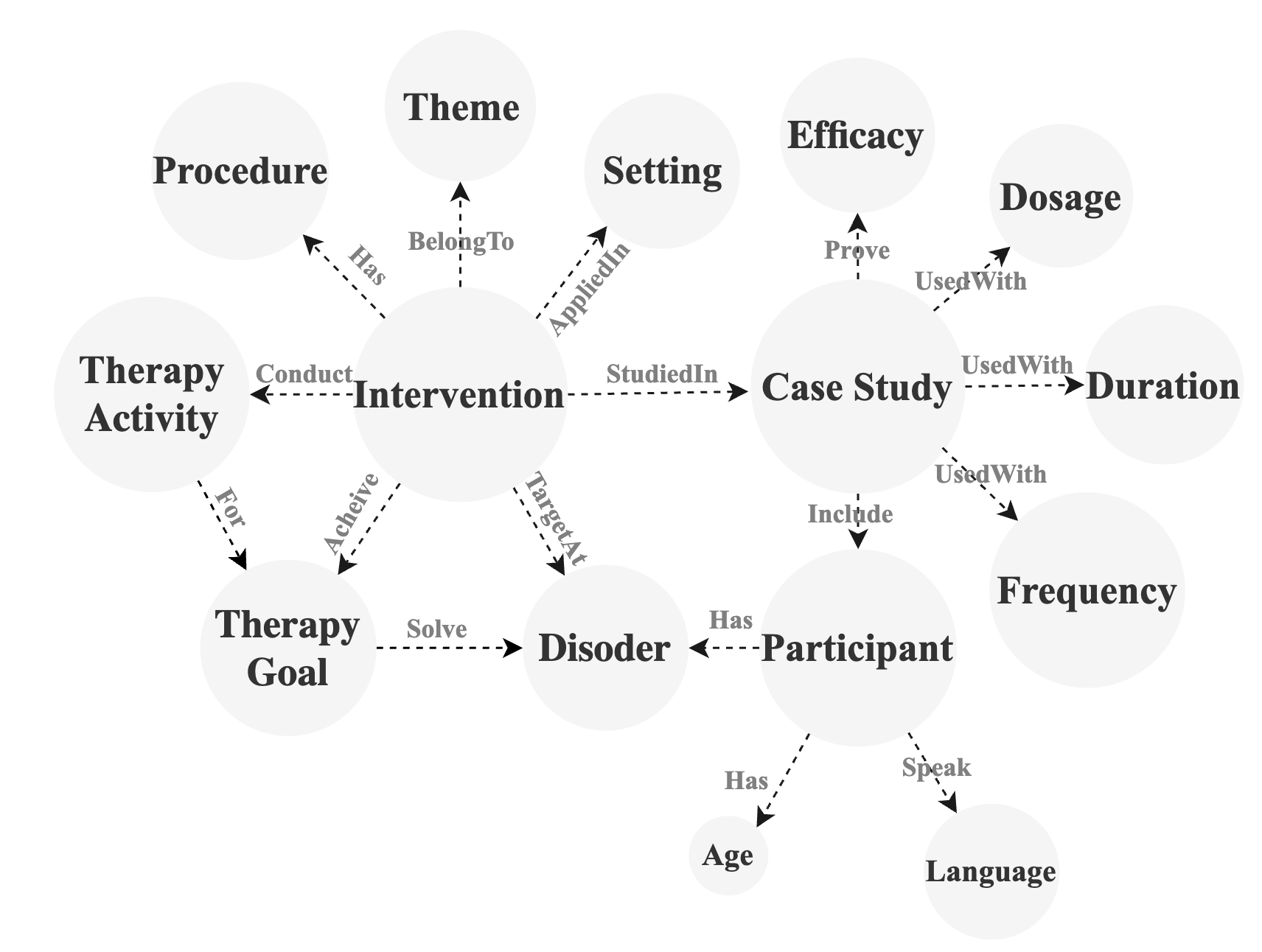}
  \caption{Ontology of speech-language intervention.}
  \vspace{-3mm}
  \label{fig:ontology}
\end{figure}
\begin{figure*}[h]
  \centering
  \begin{subfigure}[b]{0.34\textwidth}
    \centering
    \includegraphics[width=\textwidth]{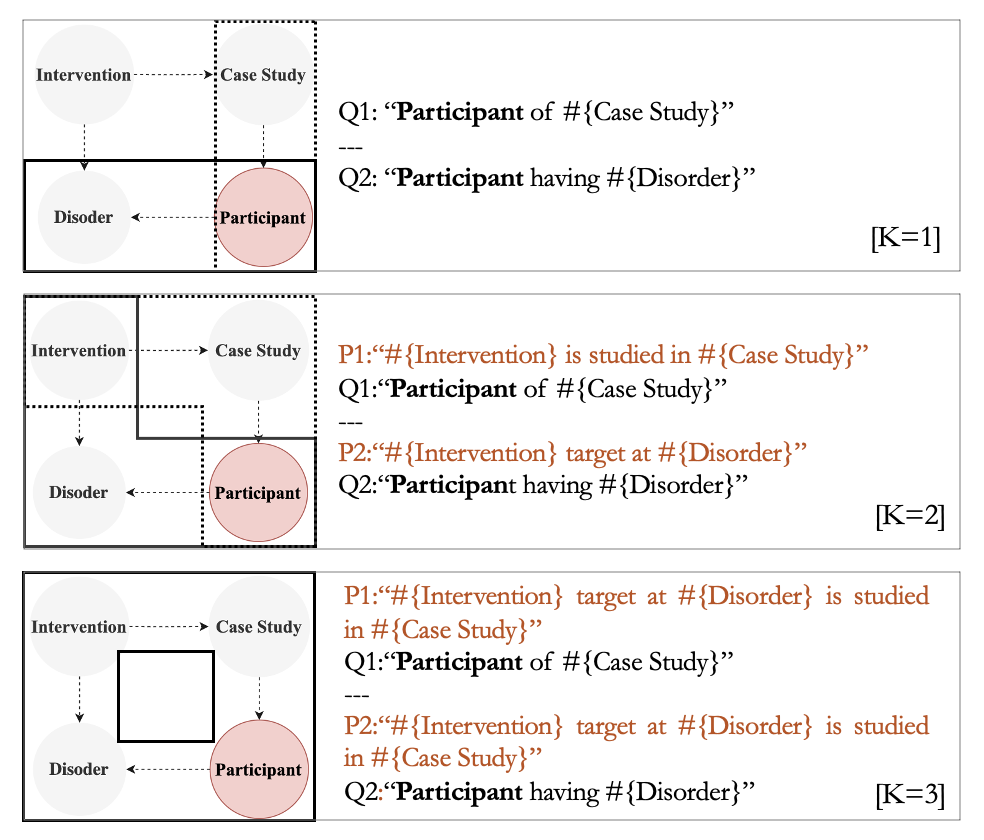}
    \caption{Illustration of `participant' annotation under different k values.}
    \label{fig:context}
  \end{subfigure}%
  \hfill
  \begin{subfigure}[b]{0.32\textwidth}
    \centering
    \includegraphics[width=\textwidth]{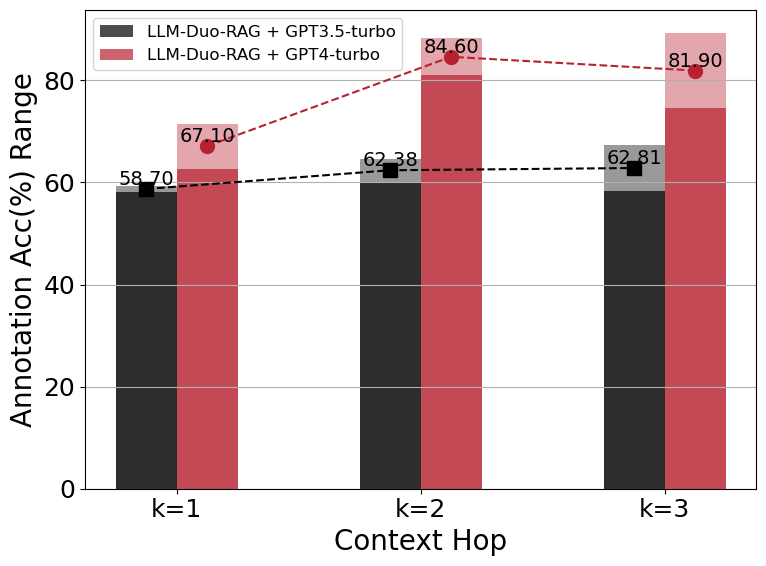}
    \caption{Range of `participant' annotation accuracy using LLM-Duo-RAG with GPTs at k=1,2,3.}
    \label{fig:kacc}
  \end{subfigure}%
  \hfill
  \begin{subfigure}[b]{0.32\textwidth}
    \centering
    \includegraphics[width=\textwidth]{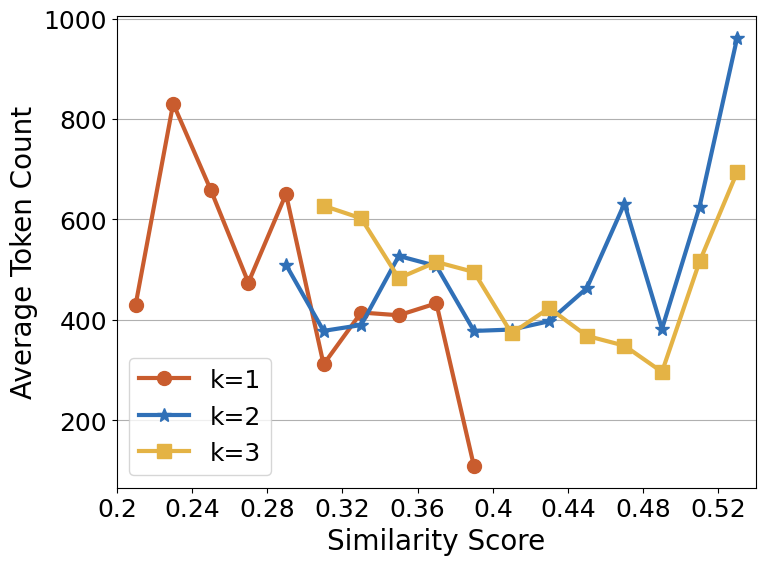}
    \caption{Token count distribution of retrieved-back chunks across varying similarity scores using `participant' annotation queries of k=1,2,3.}
    \label{fig:kscore}
  \end{subfigure}
  \caption{Evaluation of `participant' annotation with POP of different context sizes.}
  \vspace{-3mm}
\end{figure*}

\subsection{Case Study: Speech-language Intervention Discovery}
 Speech-language therapy provides interventions for individuals with speech-language deficits, enhancing their quality of life across various life stages. When choosing an intervention, evidence-based practice (EBP) is attractive as it integrates research evidence from literature into the decision-making process to ensure high-quality patient care \cite{law1996speech}. Intervention research, especially studies that offer clear intervention frameworks and comprehensive case studies, are valuable references to guide EBP designs. Intervention discovery aims to extensively gather speech-language interventions from the literature corpus as references to facilitate EBP design. It involves identifying relevant studies and extracting essential features of interventions including target disorder, procedure, efficacy, case study, therapy activity, etc., which is extremely labor-intensive for human reviewers, highlighting the efficiency of automating knowledge discovery based on LLMs. 

 To verify the effectiveness of our method in a realistic scenario, we employ our framework in a speech-language intervention discovery setting. The ontology is shown in Figure \ref{fig:ontology}. To enable a large-scale discovery, we cultivate a literature base including 64,177 papers within the domain of speech-language therapy.
\begin{table*}[h!]
  \centering
  \resizebox{1\textwidth}{!}
  {
      \begin{tabular}{ll|ccc|cccc|cc}
      \toprule 
       \textbf{Methods} & \textbf{LLM} & \multicolumn{3}{c|}{\textbf{IR}} & \multicolumn{4}{c|}{\textbf{ICA}} & \multicolumn{2}{c}{\textbf{IKC}}\\
        & & CR & ACC & Cover & CR & VI & EQ & Faith & CR & ACC  \\
        \hline
        ShortContext & \textit{GPT3.5-turbo} & - & 36.9\% & 50\% & - & 0.0249 & 5.46 & \underline{0.9667}  & - & 48.2\%  \\
        \hline
        OpenAI Assistant & \textit{GPT4-turbo} & - & 76.1\%& 69.0\% & - & 0.0631 & 4.17 & 0.7857  & - & 53.3\%  \\
        LongContext & \textit{GPT4-turbo} & - & 76.3\% & 57.1\% & - & 0.0919 & 8.64 & \textbf{1.0}  & - &  61.2\% \\
        LLM-Duo-LongContext & \textit{GPT4-turbo} & 2.17 & 81.0\% & 68.7\% & 2.5 & \underline{0.0926} & 8.68 & 0.8571 & 1.31  & 69.6\%  \\
        \hline
        RAG & \textit{GPT3.5-turbo} & - & 47.6\% & 50\% & - & 0.0319 & 7.96 & 0.8550 & -  &  48.7\%  \\
        CoT-RAG & \textit{GPT3.5-turbo} & 1.04 & 78.6\% & \underline{81\%} & 3.18 & 0.0771 & \underline{10.37} & 0.7250 & 1.07 & \underline{73.2\%} \\
        Self-Refine-RAG & \textit{GPT3.5-turbo} & 1.19 & 78.5\% & 54.4\% & 2.85 & 0.0694 & 7.17 & 0.8125  & 1.12 & 54.8\%  \\
        \hline
         \multirow{3}{*}{LLM-Duo-RAG} & \textit{GPT3.5-turbo} & 1.84 & \textbf{100\%} & \textbf{86.4\%} & 2.58& \textbf{0.1159} & \textbf{13.71} & 0.9285  &1.46 & \textbf{85.6}\%  \\
         & \textit{Llama3-instruct-70b} & 2.71 & 78.6\% & 55.6\% & 2.59& 0.0748 & 9.79 & 0.8648  & 1.52 &  61.0\% \\
        & \textit{Mistral-instruct-8x22b} & 2.30 & \underline{81.9\%} & 67.5\% & 2.16& 0.0763 & 9.87 & 0.8875  &1.46 & 67.2\%  \\      
      \bottomrule   
      \end{tabular}
  }
  \caption{Comparison of annotation results with baselines using different LLMs.}
  \vspace{-3mm}
  \label{tab:llmduocomp}
\end{table*}
\subsection{Annotation Baselines}
The core idea of our automated intervention framework is to leverage the POP algorithm for guiding the annotation process while utilizing LLM-Duo to refine initial annotations by incorporating external feedback from another LLM. Instead of setting up another LLM for evaluation, recent studies demonstrate that LLMs can engage in self-correction to enhance their responses autonomously \cite{liu2024largelanguagemodelsintrinsic}\cite{li2024confidencemattersrevisitingintrinsic}. Notable examples of this include Chain-of-Thought (CoT) \cite{wei2022chain} and Self-Refine \cite{madaan2024self}. We separately equip the explorer chatbot based on RAG with these two prompting methods for annotation and denote them as CoT-RAG and Self-Refine-RAG. Additionally, in LLM-Duo, a potential substitution of \textit{explorer} is long-context LLM, which is capable of processing entire document tokens instead of chunking and retrieval with RAG. We refer to the LLM-Duo system as LLM-Duo-RAG when using \textit{explorer} built on RAG, and as LLM-Duo-LongContext when utilizing long-context LLMs. Besides, we also compare with methods that directly input paper text to LLMs for zero-shot QA annotation without the evaluation feedback loop, including ShortContext LLM, LongContext LLM, OpenAI Assistant, and RAG. 

\subsection{Evaluation}
In the experiment of comparing LLM-Duo with annotation baselines, we report six types of metrics: 1) Consistency Rounds (CR): the number of refine loops the method makes before the annotation generation achieving consistency; 2) Verbosity Index (VI): the number of covered aspects per 1k tokens in the annotations, which is an important metric for emphasizing content completeness; 3) Enumeration Quantity (EQ): the number of items listed in the annotations (i.e., therapy activities, therapy goals.); 4) Faithfulness (Faith): the extent of the annotation faithful to the provided literature literature, which is measured by FaithfulnessEvaluator\footnote{\scriptsize{\parbox{\dimexpr\linewidth-1em}{\url{https://docs.llamaindex.ai/en/stable/examples/evaluation/faithfulness_eval}}}} of Llamaindex. 5) Accuracy (ACC): the percentage of correct annotations in all LLM-generated annotations. 6) Cover: the percentage of correct LLM-generated annotations to the total mentioned concept entities in the provided literature. 

\section{Results}
In this section, we first provide a detailed evaluation of our progressive ontology prompting algorithm and the LLM-Duo annotation framework. Then, we showcase the results of speech-language intervention discoveries using our automated intervention discovery framework.

\subsection{POP Algorithm Study}
\noindent\textbf{Context Size Analysis.} In the POP algorithm, the context size $k$ determines the diversity and volume of information included in the intervention annotation prompt. To assess the impact of context size on annotation quality, we conducted experiments using various $k$ values to generate intervention annotation prompts for LLM-Duo-RAG. As shown in Figure~\ref{fig:context}, we annotate the `participant' concept for the experiment, which was based on a random selection of 8 speech-language pathology literature. 

The annotation accuracy is shown in Figure~\ref{fig:kacc}. The results indicate that as the context size $k$ increases, annotation accuracy improves significantly, suggesting that a larger context provides more informative prompts, thereby enhancing annotation quality. Moreover, GPT-4-turbo consistently outperforms GPT-3.5-turbo across all $k$ values, demonstrating that more advanced language models can further improve annotation accuracy. Besides, we inspect the text chunks retrieved back by different informative annotation queries based on various $k$ values. We report the range of similarity scores and token count distribution of retrieved-back chunks in Figure~\ref{fig:kscore}. The similarity score represents the semantically relevancy between retrieved texts to annotation queries. The results show that for $k=1$, the retrieved text chunks generally have low similarity to the query, and the token count decreases as the similarity score increases, leading to lower annotation quality. In contrast, higher $k$ values, especially $k=2$, yield more relevant retrievals. For $k=2$, the token count increases with higher similarity scores, indicating that richer and more relevant content is captured, resulting in improved annotation quality.

\noindent\textbf{Prioritized BFS Analysis.}
In the POP algorithm, we employ the out-to-in ratio to prioritize neighboring nodes during BFS-based prompt creation and scheduling. This strategy ensures that nodes with more outgoing edges are visited first, allowing them to provide more context for annotating other nodes. For example, one annotation sequence following prioritized POP over the speech-language intervention ontology is `TherapyActivity'$\rightarrow$`TherapyGoal'$\rightarrow$`Disorder'. In this section, we compare the `Disorder' annotation results using the POP algorithm with and without prioritization, as well as one-shot annotation without using POP, where the entire KG schema is included on a single annotation prompt to extract all triplets. The results are presented in Table~\ref{tab:wwo}. We can observe that applying both the POP and prioritized BFS notably enhances annotation performance.
\begin{table}[H]
  \centering
  \resizebox{1\columnwidth}{!}
  {
  \begin{tabular}{lcc}
    \toprule
    \textbf{LLM-Duo-RAG} & \textbf{GPT3.5-turbo} & \textbf{GPT4-turbo} \\
    \hline
    POP\textcolor{red}{\ding{55}} & 68.18 & 72.73 \\
    POP\textcolor{green}{\ding{51}} Prioritized-BFS\textcolor{red}{\ding{55}} & 77.28 & 83.20 \\
    POP\textcolor{green}{\ding{51}} Prioritized-BFS\textcolor{green}{\ding{51}} & \textbf{81.82} & \textbf{86.37} \\
    \bottomrule
  \end{tabular}
}
  \caption{Comparison of annotation results with and without the POP and prioritized BFS.}
  \label{tab:wwo}
\end{table}
\begin{figure*}[h]
  \centering
  \begin{subfigure}[b]{0.25\textwidth}
    \centering
    \includegraphics[width=\textwidth]{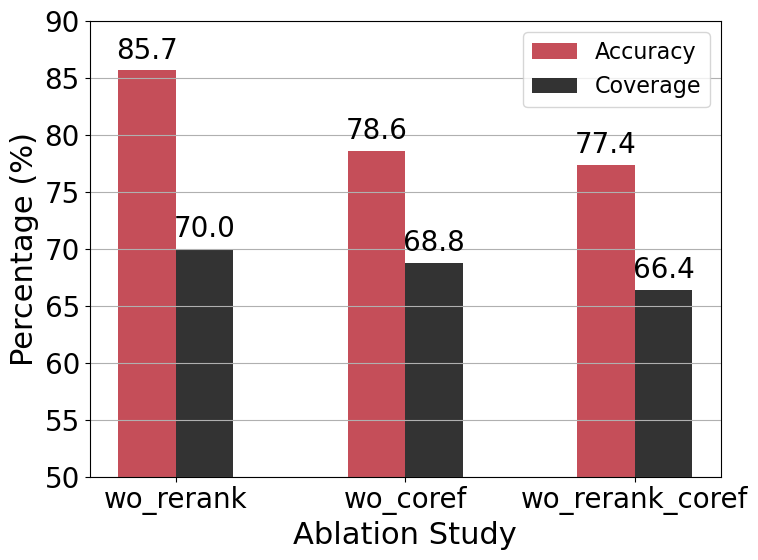}
    \caption{Ablation studies of removing rerank (wo\_rerank) and corefenence (wo\_coref) modules in LLM-Duo-RAG.}
    \label{fig:ablation}
  \end{subfigure}%
  \hfill
  \begin{subfigure}[b]{0.38\textwidth}
    \centering
    \includegraphics[width=\textwidth]{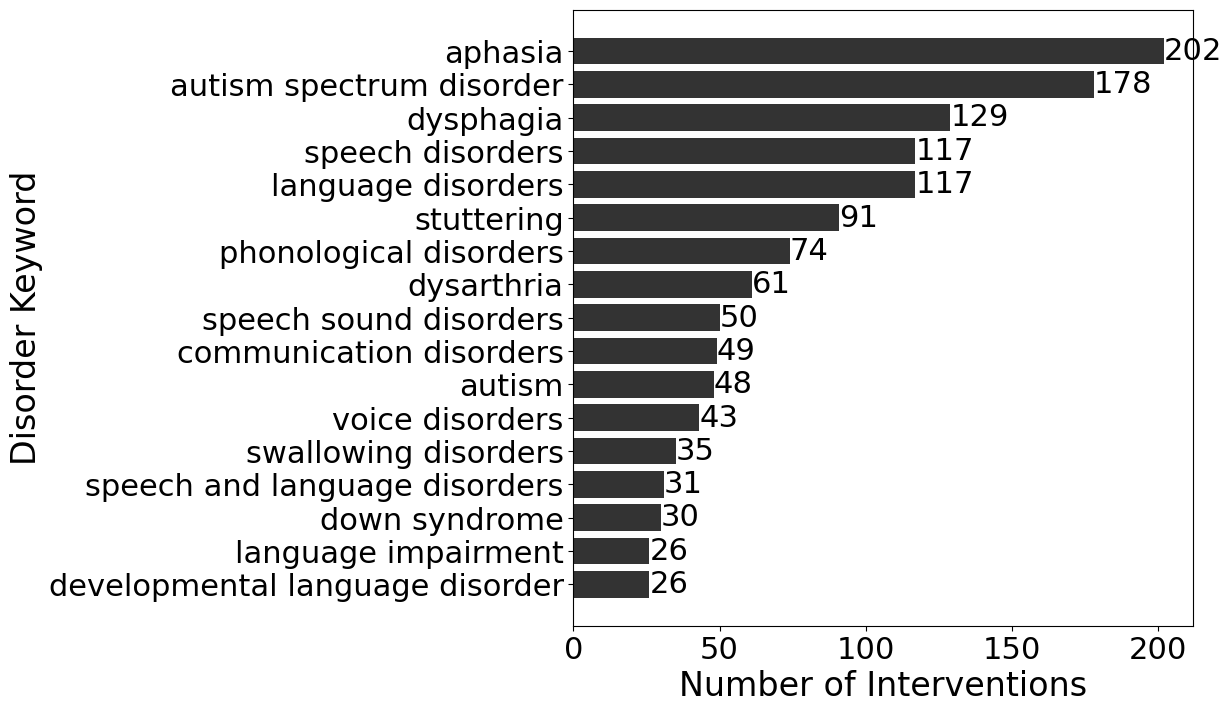}
    \caption{Number of discovered interventions for top 20 disorders. }
    \label{fig:disorderiv}
  \end{subfigure}%
  \hfill
  \begin{subfigure}[b]{0.31\textwidth}
    \centering
    \includegraphics[width=\textwidth]{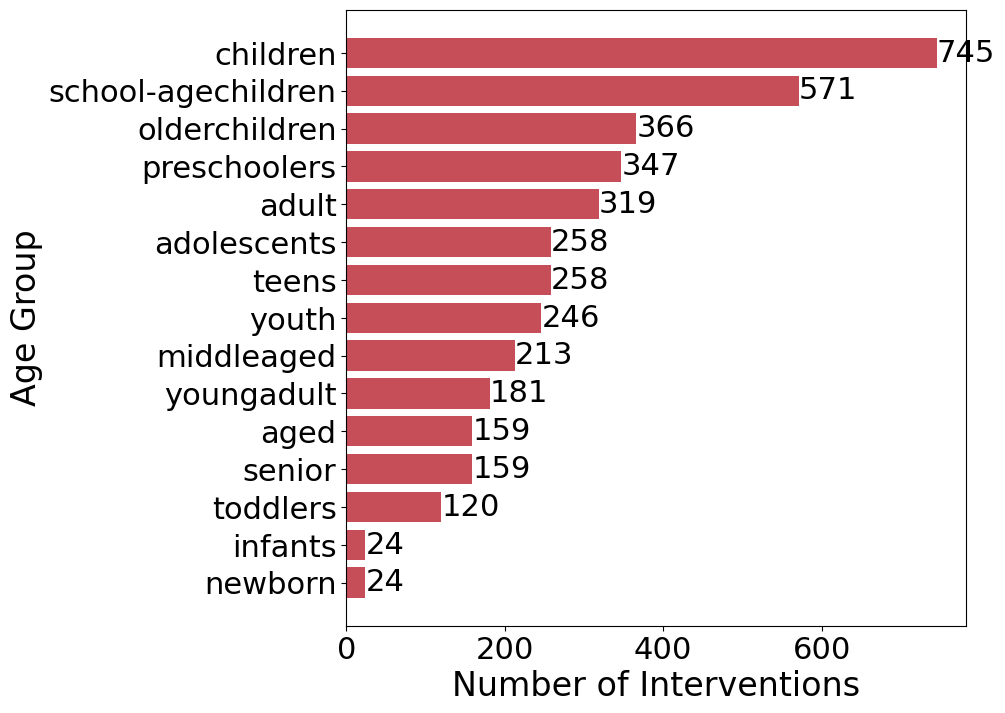}
    \caption{Number of discovered interventions for different age groups.}
    \label{fig:agegroupiv}
  \end{subfigure}
  \caption{Ablation study and annotated speech-language intervention statistics.}
  \vspace{-5mm}
\end{figure*}
\begin{table*}[h]
  \centering
  \resizebox{\textwidth}{!}{
    \begin{tabular}{m{4cm}m{2cm}p{20cm}}
      \toprule
      \textbf{Disorder} & \textbf{Interventions} & \textbf{Intervention Examples}\\
      \midrule
      Aphasia & 202 & Phonological therapy, Semantic therapy, Syntax Stimulation Program, Melodic Intonation Therapy (MIT), Multimodal Speech Therapy with tDCS, Cross-Language Generalization Therapy (CLGT), Word Learning Paradigm \\
      \midrule
      Autism Spectrum Disorder & 178 & Personalized Idiom Intervention (PII), Classroom-wide peer tutoring, Idiom Isolation Intervention, Hanen More Than Words, Parent-Mediated Communication-Focused Treatment, Picture Exchange Communication System (PECS) \\
      \midrule
      Dysphagia & 129 & Swallowing Maneuver Therapy, Focal Vibration Therapy (FVT),  Oral Neuromuscular Training and Vibrational Therapy, Prophylactic Swallowing Intervention, High-speed jaw-opening exercise, Palatal Augmentation Prostheses (PAP) \\
      \midrule
      Stuttering & 91 & Lidcombe Program, Syllable-timed speech, Electronic devices for stuttering, Computer software for stuttering, Bone Conduction Delayed Feedback Therapy, Fluency Techniques and Fear Reduction, Cognitive Behavioral Therapy (CBT) \\
      \midrule
      Phonological Disorder & 74 & Nonlinear Phonological Intervention Program, Metronome-paced Speech Therapy, Phonological Awareness and Articulatory Training (PAAT), Phonological Meaning Therapy, Motor-based Intervention Approach\\
      \midrule
      \textbf{AgeGroup} & \textbf{Interventions} & \textbf{Intervention Examples}\\
      \midrule
      Children & 745 & Pharyngeal flap procedure, National Health Service (NHS) 1-week intensive course, Ultrasound Visual Biofeedback (U-VBF), Intensive Speech Therapy, Community-Based Speech Therapy Models, Early Vocal Intervention, Auditory-Verbal Therapy (AVT) \\
      \midrule
      School-age Children & 571 & Intensive Speech Therapy, Early Vocal Intervention, APD intervention, Auditory-Verbal Therapy (AVT), Multisensory Stimulation Therapy, Oral Functional Training (OFT), Rhythmic Reading Training (RRT), Rapid Syllable Transition Treatment (ReST) \\
      \midrule
      Older children & 366 & Semantic Categorization Therapy, Early Vocal Intervention, Rhythmic Reading Training (RRT), Speech Bulb Reduction Program, Intensive speech and language therapy, Peer-Mediated Intervention, Lidcombe Program, Oral Functional Training (OFT) \\
      \midrule
      Preschoolers & 347 & Early Vocal Intervention, Treatment-as-usual, The Lidcombe Program, Oral Functional Training (OFT), Speech Production Therapy with Reward System, Phonological Interventions and Contrast Therapy, Cycles Phonological Remediation Approach \\
      \midrule
      Adult & 319 & Pharyngeal flap procedure, Linguistic Retrieval Therapy (LRT), Oral Hydration Intervention, Physiologic Swallowing Therapy, Myofunctional Intervention (OMT), Orthognathic speech therapy, Eye-Tongue Movement Training, Behavioral Voice Therapy\\
      \bottomrule
    \end{tabular}
  }
  \caption{Intervention-disorder examples in our discoveries.}
  \label{tab:Examples}
  \vspace{-5mm}
\end{table*}
\subsection{LLM-Duo with Baselines}
In this section, we compare LLM-Duo with several advanced baselines using annotations of 8 randomly selected papers from our speech-language literature corpus. The evaluation focuses on three key dimensions: 1) Intervention Recognition (IR), identifying intervention entities within the literature; 2) Intervention Aspect Summary (IAS), annotating the key aspects (e.g., procedure, therapy activity, therapy goals) of the intervention by capturing and summarizing relevant information from the paper; and 3) Intervention Knowledge Completion (IKC), linking interventions to theme classes (e.g., speech awareness, speech articulation, comprehension, foundation skills, etc.) and setting concept nodes (e.g., home, healthcare facilities, schools, teletherapy, etc.). We use human annotators for the IR and IKC tasks to generate gold-standard results for comparison. In the IAS task, we only ask human annotators to tag relevant text fragments related to specific intervention aspects due to potential individual bias in human interpretation.

The experimental results are reported in Table~\ref{tab:llmduocomp}. It should be noted that we implemented `ShortContext' using Llama3-instruct-70b (FP16) and Mistral-instruct-8x22b models (INT8). However, directly prompting these models with full paper text fails to produce annotations in a zero-shot QA setting. Their generations do not align with the annotation questions. The results in Table~\ref{tab:llmduocomp} show that LLM-Duo-RAG outperforms all baselines. Although GPT4-turbo has a 128k context window length and is capable of generating annotations, its annotation coverage remains inadequate. Integrating it with the LLM-Duo framework can significantly improve both the accuracy and the comprehensiveness of the intervention annotations. Additionally, compared with simple RAG, self-correct prompting methods such as CoT and Self-Refine can significantly enhance intervention annotations, but their performance is still worse than LLM-Duo-RAG. Instead of utilizing costly GPT models, LLM-Duo-RAG, which employs open-source models including Llama3-instruct-70b and Mistral-instruct-8x22b, can achieve comparable annotation quality.

\subsection{Ablation Study}
In our implementation, the RAG technique serves as the backbone of \textit{explorer}. We employ `FastCoref' for coreference resolution and rerank retrieved chunks by similarity score using the `cross-encoder/ms-marco-MiniLM-L-2-v2 model'. This section presents ablation studies for both components. We report the accuracy of intervention recognition in this section. As shown in Figure~\ref{fig:ablation}, the results demonstrate that removing these components significantly decreases annotation accuracy, showing the necessity of each module.

\subsection{Speech-Language Intervention Discovery}
Through our automated intervention discovery framework, we identified 2,421 interventions supported by case studies from 64,177 literature in the speech-language pathology domain. The statistics of discovered interventions are presented in Figure~\ref{fig:disorderiv} and Figure \ref{fig:agegroupiv}. More intervention examples are provided in Table~\ref{tab:Examples}. 19 clinicians and students reviewed our annotations through online Google forms. We have constructed the first intervention knowledge graph in the speech-language pathology domain, which will be made publicly accessible upon acceptance. This knowledge graph is anticipated to be a valuable resource for domain experts, facilitating evidence-based clinical decision-making, question-answering, and recommendation systems, ultimately contributing to improved healthcare outcomes.

\section{Conclusion}
In this paper, we developed a novel LLM-based framework for automatic intervention discovery from literature, featuring a progressive ontology prompting algorithm and a dual-agent system.
The proposed method achieves superior performance compared with advanced baselines, enabling more accurate intervention discoveries. Our approach is adaptable to various intervention ontologies in healthcare and offers practical value to improve healthcare quality.
\section*{Acknowledgments}
The research was supported, in part, by the National AI Institute for Exceptional Education (NSF Award \#2229873), Center for Early Literacy and Responsible AI (IES Award \#R305C240046), FuSe-TG (NSF Award \#2235364) and SaTC (NSF Award \#2329704). The opinions expressed are those of the authors and do not represent the views of any sponsors.
\bibliographystyle{named}
\bibliography{ijcai25}
\clearpage
\appendix
\section{POP Algorithm}
\subsection{Algorithm Pseudocode}
\label{sec:A1}
\begin{algorithm}
\scriptsize 
\caption{Progressive Ontology Prompting}\label{alg:pop}
\begin{algorithmic}[0] 
\Function{OntologyTraversal}{$\mathcal{G}$, $k$}
    \State Initialize: $queue, Order \gets []$, $visited, Khops \gets \{\}$
    \State Identify the source node $s$
    \State $enqueue(queue, (s, 0))$, $visited[s] \gets \text{true}$, $Khops[s] \gets []$
    \While{not $queue$.empty()}
        \State $(current, hop) \gets dequeue(queue)$
        \State $Order$.append$(current)$
        \State Gather unvisited neighbors of $current$ into $neighbors$
        \State Sort $neighbors$ by out-to-in ratio in descending order
        \For{$neighbor$ in $neighbors$}
            \If{not $visited[neighbor]$}
                \State $enqueue(queue, (neighbor, hop + 1))$
                \State $visited[neighbor] \gets \text{true}$
                \State $Khops[neighbor] \gets []$
            \EndIf
            \If{$hop + 1 \leq k$}
                \State $Khops[neighbor].append(current)$ 
            \EndIf
        \EndFor
    \EndWhile
    \State \Return $Order$, $Khops$
\EndFunction

\Function{Main}{$\mathcal{G}$, $k$, $T$}
    \State Initialize $annotations \gets \{\}$
    \State $Order, Khops \gets \Call{OntologyTraversal}{\mathcal{G}, k}$
    \While{not $Order$.empty()}
        \State $current \gets Order$.dequeue(), $context \gets Khops[current]$
        \If{not $context$.empty()}
            \State $discoveries \gets \{annotations[t] \text{ for } t \text{ in } context\}$
            \State $prompt \gets T(current, G_{sub}(context), discoveries)$
        \Else
            \State $prompt \gets T(current)$
        \EndIf
        \State $annotations[current] \gets \Call{LLM}{prompt}$
    \EndWhile
\EndFunction
\end{algorithmic}
\end{algorithm}

\subsection{Annotation Prompt Template Generation}
\label{sec:A2}
In the POP algorithm, the annotation prompt for a specific concept node is generated from a template $T$, which is constructed using the node's context (the k-hop visited neighborhood) and the completed annotations within that context. We carefully design the prompt to task LLM to generate $T$ as follows: \\
\\
{\ttfamily{\small
The following triplets outline an annotation ontology:\{(Intervention, StudiedIn, Case Study), (Intervention, Include, Participant), (Case Study, UsedWith, Frequency)\}. All concept nodes have been annotated except from \{\#Frequency\}. Your task is to create all possible annotation prompt templates for the \{\#Frequency\}, leveraging the ontology structure. \\
Example: \\
Ontology: [(Intervention, StudiedIn, Case Study), (Intervention, TargetAt, Disorder), (Case Study, Include, Participant), (Participant, Has, Disorder)]. \\
Annotate: Participant \\
Prompts: \\
T1: Who is the participant in \{\#case study\}? \\
T2: \{\#Intervention\} is studied in \{\#case study\}. Who is the participant in \{\#case study\}? \\
T3: \{\#Intervention\}, targeting \{\#Disorder\}, is studied in \{\#case study\}. Who is the participant in \{\#case study\}? \\}}
\\
Some example output of GPT3.5 for the above prompt is: \\
\\
{\ttfamily{\small
T1: What is the frequency used in \{\#case study\}? \\
T2: In the context of \{\#intervention\} studied in \{\#case study\}, what is the frequency of the intervention? \\
T3: \{\#Intervention\} is studied in \{\#case study\} with \{\#Participant\}. What is the frequency of the intervention in \{\#case study\}? \\}}
\\
Since template $T$ is composed of concept nodes visited before the current node, the final annotation prompt for the current node is derived by incorporating the annotations associated with these concept nodes into $T$.
\section{Speech-Language Intervention Annotation}
\label{sec:B}
\begin{table*}
  \resizebox{\textwidth}{!}
  {
      \begin{tabular}{lp{24cm}}
        \toprule
        Theme&Definition\\
        \midrule
        speech awareness & work that involves recognizing and understanding speech sounds. It includes phonological awareness (identifying and manipulating sounds), auditory discrimination (distinguishing sounds), and sound identification (recognizing the meaning of sounds).\\
        speech articulation & work that refers to the physical production of speech sounds using the mouth, lips, tongue, and respiratory system. It focuses on the clarity and accuracy of pronouncing phonemes and forming them into words. \\
        comprehension & work that aims to improve the understanding of (receptive) language. \\
        expressive language & work that aims to improve the children's expressive language, in quantity, vocabulary or structure. \\
        self-monitoring & work designed to help the patient's awareness of their speech and language difficulties and how they might be able to overcome them. \\
        generalisation & work to help make speech and language or therapy gains transferable to other situations and environments. \\
        foundation skills & work to practise and improve a range of early skills, many of which might be considered foundations for speech and language development. \\
        functional communication & work focusing on those aspects of communication that help the child's involvement and participation in life situations; this might be functional language, signing or the use of symbols. \\
        adult understanding and empowerment & work that helps parents to understand the nature of their child's speech and language difficulty, what helps to improve it and why. \\
        adult-child interaction & work on the interaction between the parent/adult and the child. All of the changes to adult/parent-child interactions were emphasised in terms of those that encourage speech and language development. \\
        
      \bottomrule
    \end{tabular}
    }
\caption{Definition of intervention themes.}
\label{tab:theme}
\end{table*}
\subsection{Literature Corpus}
We cultivate a literature corpus of 64,177 research articles within the domain of speech-language pathology to facilitate intervention discovery. To conduct our literature search, we use a collection of carefully selected keywords drawn from a glossary of commonly used terms in speech-language pathology. These keywords include: \\
\\
\textit{``speech language therapy, speech language disorder, speech sound disorder, articulation disorder, speech intervention, language intervention, auditory discrimination, auditory processing disorder, phonological awareness, phonological processes, auditory perception, babbling, motor speech disorder, morpheme, phonology, prosody, stuttering, language impairment, speech language pathologist, speech and language therapist, babbling, expressive language delay, cleft speech disorder, autism spectrum disorder, developmental phonological disorder, developmental stuttering, phonological impairment, developmental dysarthria, down syndrome, swallowing disorder, communication impairment, articulation impairment, dyslexia, apraxia, dysarthria, dysphagia, communication disorder, expressive language disorder, dyspraxia, aphasia, augmentative and alternative communication, central auditory processing disorder, cleft lip and palate, down syndrome, fluency disorders, hearing loss, orofacial myofunctional disorders, spoken language disorders, written language disorders, acquired brain injury, apraxia of speech, auditory comprehension, literacy impairments, voice difficulties, language-based learning disabilities."}
\subsection{Annotation Ontology}
Speech-language pathology provides interventions for individuals with speech-language deficits, improving their quality of life at various stages. When choosing an intervention, evidence-based practice (EBP) is attractive as it integrates research evidence from literature into the decision-making process to ensure high-quality patient care \cite{law1996speech}. Research on interventions, especially those presenting clear frameworks and comprehensive case studies, serves as valuable guidance for designing EBPs.  In this paper, we apply our automated framework to speech-language intervention discovery, with the intervention ontology shown in Figure~\ref{fig:ontology}. A detailed explanation of the concepts within the ontology is provided below:
\begin{itemize}
    \item \par\textbf{Intervention} represents a targeted treatment practice designed to enhance an individual’s communication skills.
    \item \par\textbf{Disorder} represents the type of disorder that causes difficulties in an individual's voice, speech, language, or swallowing functions.
    \item \par\textbf{Setting} represents a specific environment where interventions are implemented. We identify six key settings: home, healthcare facilities (such as hospitals or rehabilitation centers), early childhood centers (like nurseries or daycare), schools, clinics and private practices, and teletherapy.
    \item \par\textbf{Theme} represents the theme of the intervention. As shown in Table~\ref{tab:theme}, we categorize interventions into 10 themes based on their characteristics and therapy goals. 
    \item \par\textbf{Therapy Activity} represents a task designed to address a particular speech or language challenge in an individual, such as using a minimal pairs activity to enhance phonological awareness.
    \item \par\textbf{Therapy Goal} represents a specific area that the intervention is designed to enhance.
    \item \par\textbf{Procedure} represents a comprehensive description of how the intervention is carried out.
    \item \par\textbf{Efficacy} represents the conclusion about the effectiveness of the intervention.
    \item \par\textbf{Frequency/Dosage/Duration} represents the frequency/dosage/ duration of the intervention practiced in the case study that demonstrates its efficacy. 
    \item \par\textbf{Case Study} represents a detailed examination of the intervention on a particular individual or group with communication disorders. The purpose of a case study is to provide a deep understanding of the patient's unique needs and assess the intervention's effectiveness.
    \item \par\textbf{Participant} represents the individuals or populations that are involved in the case study of intervention.
    \item \par\textbf{Age} represent the age of experiment participant
or claimed target population of the intervention. The age is quantified with a granularity of half a year. We additionally convert age to age groups including \textit{``newborn, infants, toddlers, children, preschoolers, school-age children, older children, youth, teens, adolescents, adult, young adult, middle aged, aged, senior"}. Each specific age may be associated with multiple age groups. For instance, an individual aged 13 years could be categorized into the `teens,' `adolescents,' and `children' age groups.
    \item \par\textbf{Language} represents the speaking language of the experiment participant in the case study of the intervention. 
\end{itemize}

\section{CoT \& Self-Refine Baselines}
\label{sec:C}
Instead of using an external LLM to provide evaluation feedback to the explorer, we use CoT and SelfRefine prompting techniques as baselines to task explorer refine annotations independently. The prompts of CoT and SelfRefine are as follows: \\
\begin{lstlisting}[language=json]
Background: Your last answer to my question {#init annotation question} is: {#last annotation}.
---
*CoT:
    Prompt_CoT1: "{#Background} Make a plan to correctly answer my question again."
    Prompt_CoT2: "{#Background} Make a plan to answer my question again with more comprehensive results."
---
*SelfRefine:
    Prompt_SelfRefine_Feedback1: "{#Background} Reflect your answer. Analyze the correctness of the information provided. Provide critque to help improve the answer. Your feedback:"
    Prompt_SelfRefine_Refine1: "{#Background} Critics: {#feedback}. Based on your last answer and its critics, revise your answer to my question. Your answer:"
    Prompt_SelfRefine_Feedback2: "{#Background} Reflect your answer. Analyze the included aspects in your answer. Provide critque to help make the response more comprehensive. Your feedback:"
    Prompt_SelfRefine_Refine2: "{#Background} Critics: {#feedback}. Based on your last answer and its critics, revise your answer to my question. Your answer:"
\end{lstlisting}
\section{Annotation Example}
\label{sec:D}
Below is an example of an intervention annotation from the paper "Intensive Treatment of Dysarthria Secondary to Stroke" \cite{mahler2012intensive}. \\
\begin{lstlisting}[language=json]
"Intervention": "Lee Silverman Voice Treatment (LSVT LOUD)",
"Disorder": [
        "stroke",
        "ataxia",
        "multiple sclerosis",
        "traumatic brain injury (TBI)"
],
"Procedure": "The therapy process of conducting the Lee Silverman Voice Treatment (LSVT LOUD) intervention involves intensive high-effort exercises aimed at increasing vocal loudness to a level within normal limits using healthy and efficient voice techniques. The treatment protocol includes sessions four times a week for 4 weeks, totaling 16 individual one-hour sessions. Each session consists of tasks such as maximal sustained vowel phonation, pitch range exercises, and reading functional phrases at individual target loudness levels. The second half of each session progresses to functional speech tasks, moving from words and phrases to conversation over the course of the 16 sessions. Additionally, participants are assigned daily homework to practice using normal loudness and facilitate generalization of normal loudness outside the treatment room.",
"Frequency": "four times a week",
"Dosage": "one-hour session.",
"Duration": "4 weeks, totaling 16 individual one-hour sessions.",
"Efficacy": "The outcome of the Lee Silverman Voice Treatment (LSVT LOUD) intervention in this study showed positive changes in acoustic variables of speech for all participants with dysarthria secondary to stroke. There were statistically significant increases in vocal dB SPL for sustained vowel phonation and speech tasks, indicating improvements in loudness levels and phonatory stability. Additionally, post-treatment speech samples were rated as having better voice quality and sounding more natural, suggesting an amelioration of dysarthria characteristics. Participants also reported increased confidence in their speech during post-treatment interviews.",
"Therapy Goal": [
        "Increase vocal loudness to a level within normal limits",
        "Use healthy and efficient voice techniques",
        "Progress from words and phrases to conversation over 16 sessions",
        "Facilitate generalization of normal loudness outside the treatment room"
],
"Participant": "Four participants (P01 to P04).",
"Age": "Participants in the study ranged in age from 50 to 74 years.",
"Language": "assume to be English.",
"Case Study": "Case studies and experiments regarding the Lee Silverman Voice Treatment (LSVT LOUD) intervention in the paper include studies on people diagnosed with stroke, ataxia, multiple sclerosis, and traumatic brain injury (TBI). These studies have shown improvement in articulatory features as well as loudness.",
"Therapy Activity": [
        "Maximal sustained vowel phonation",
        "Pitch range exercises",
        "Reading 10 functional phrases at individual target loudness levels",
        "Functional speech tasks progressing from words and phrases to conversation"
],
"Setting": "home",
"Theme": "speech articulation"
\end{lstlisting}
\section{Speech-Language Intervention Knowledge Base}
\label{sec:E}
\begin{figure*}[h]
    \centering
    \includegraphics[width=\textwidth]{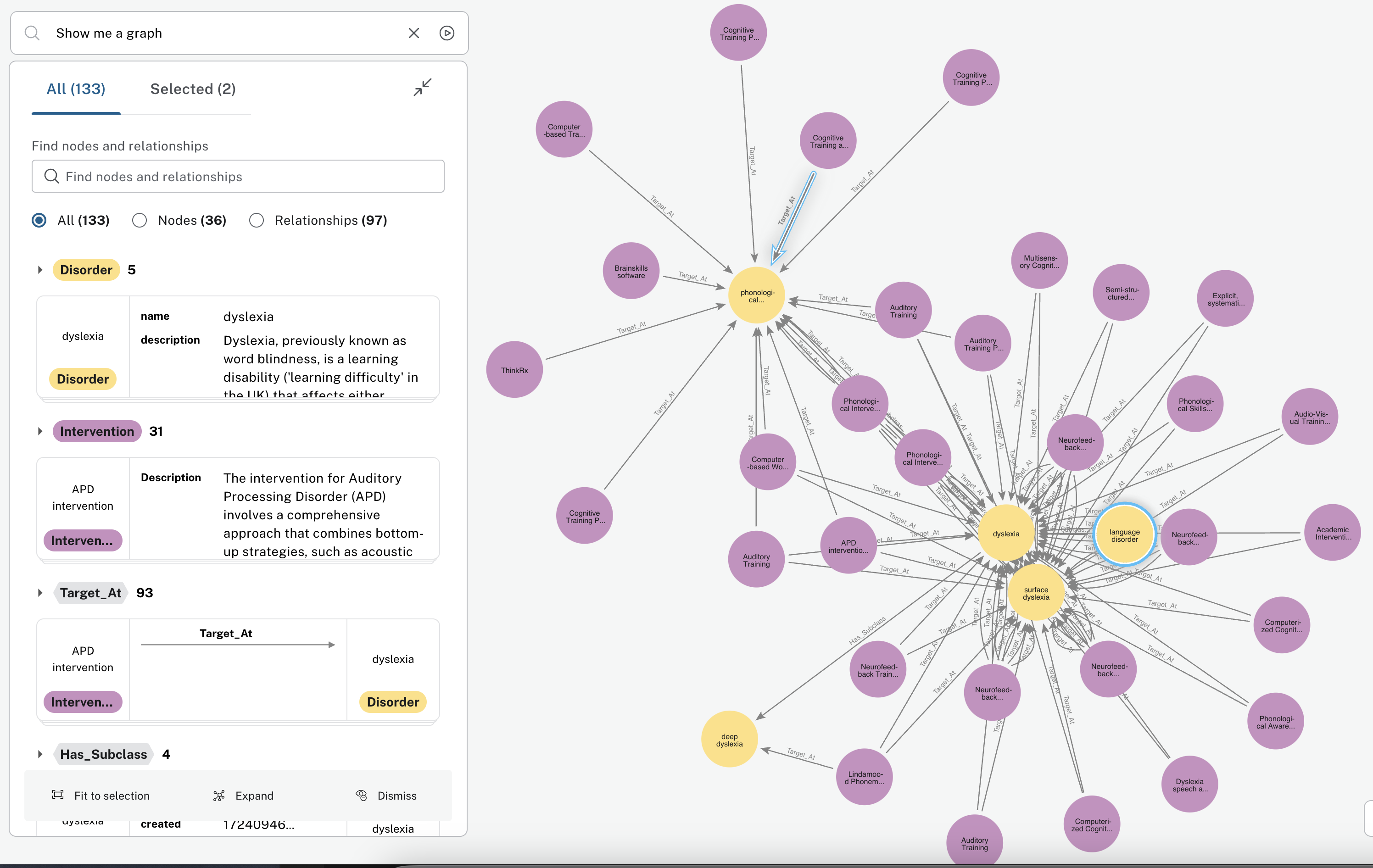}
    \includegraphics[width=\textwidth]{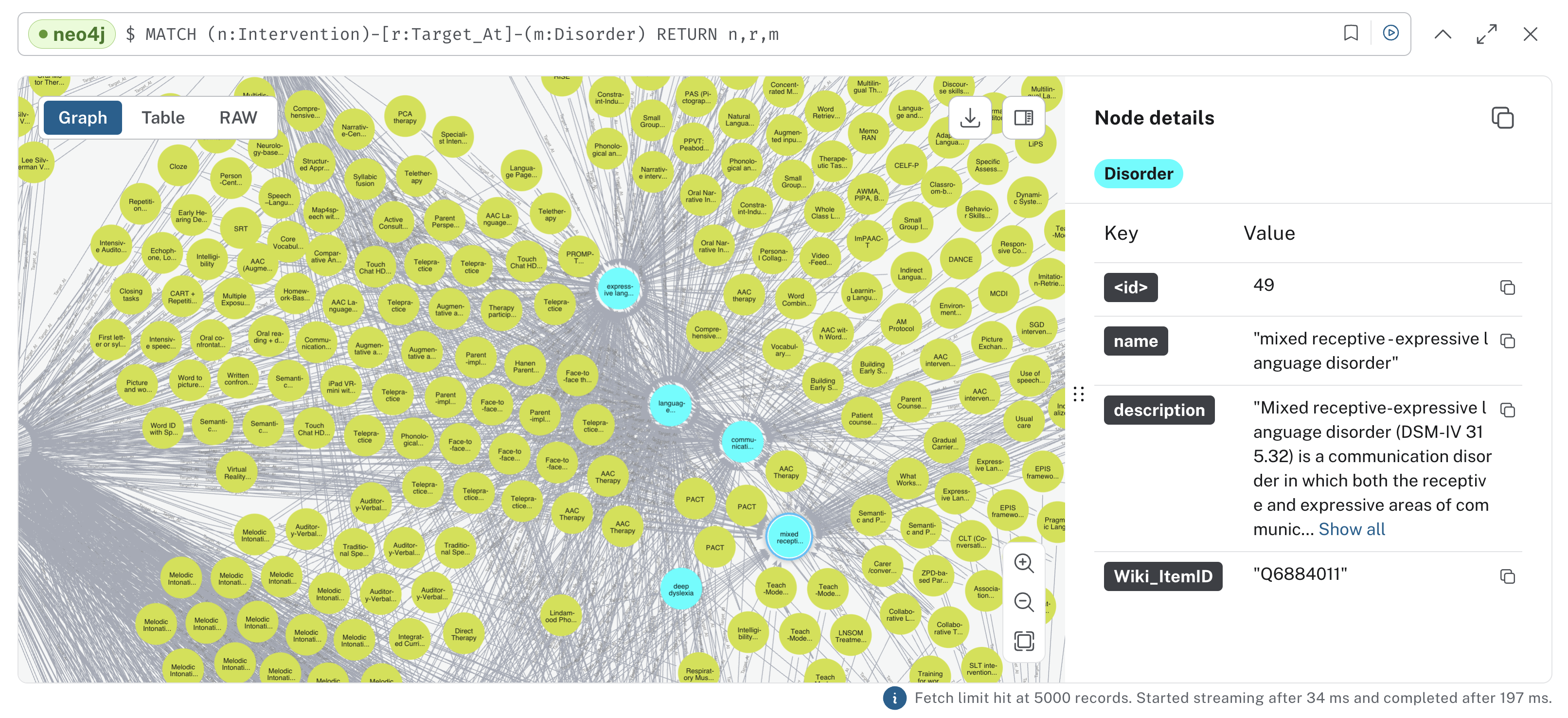}
    \caption{Partial views of Intervention-Disorder in the intervention KG.}
    \label{fig:kg1}
\end{figure*}

\begin{figure*}[h]
    \centering
    \includegraphics[width=\textwidth]{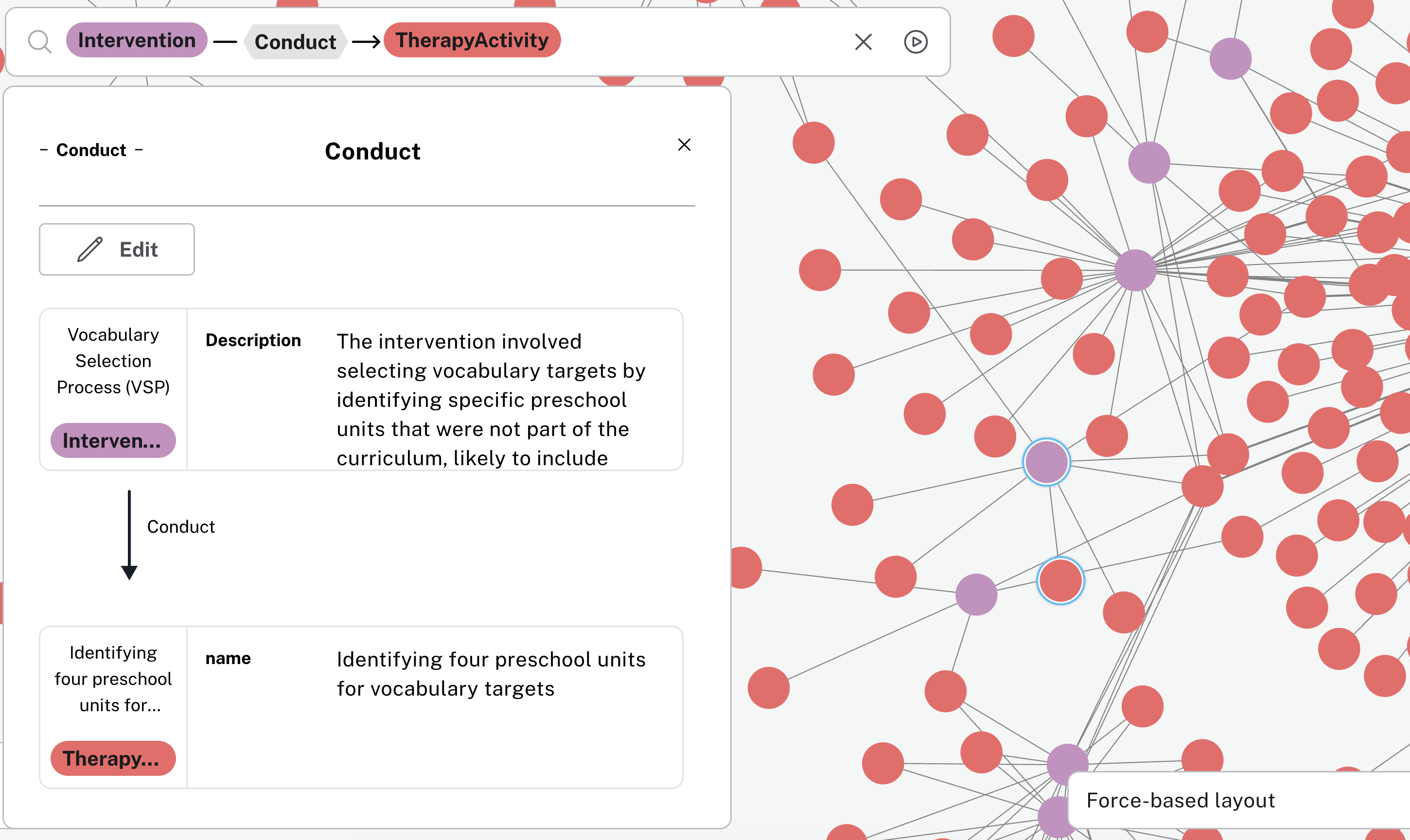}
    \includegraphics[width=\textwidth]{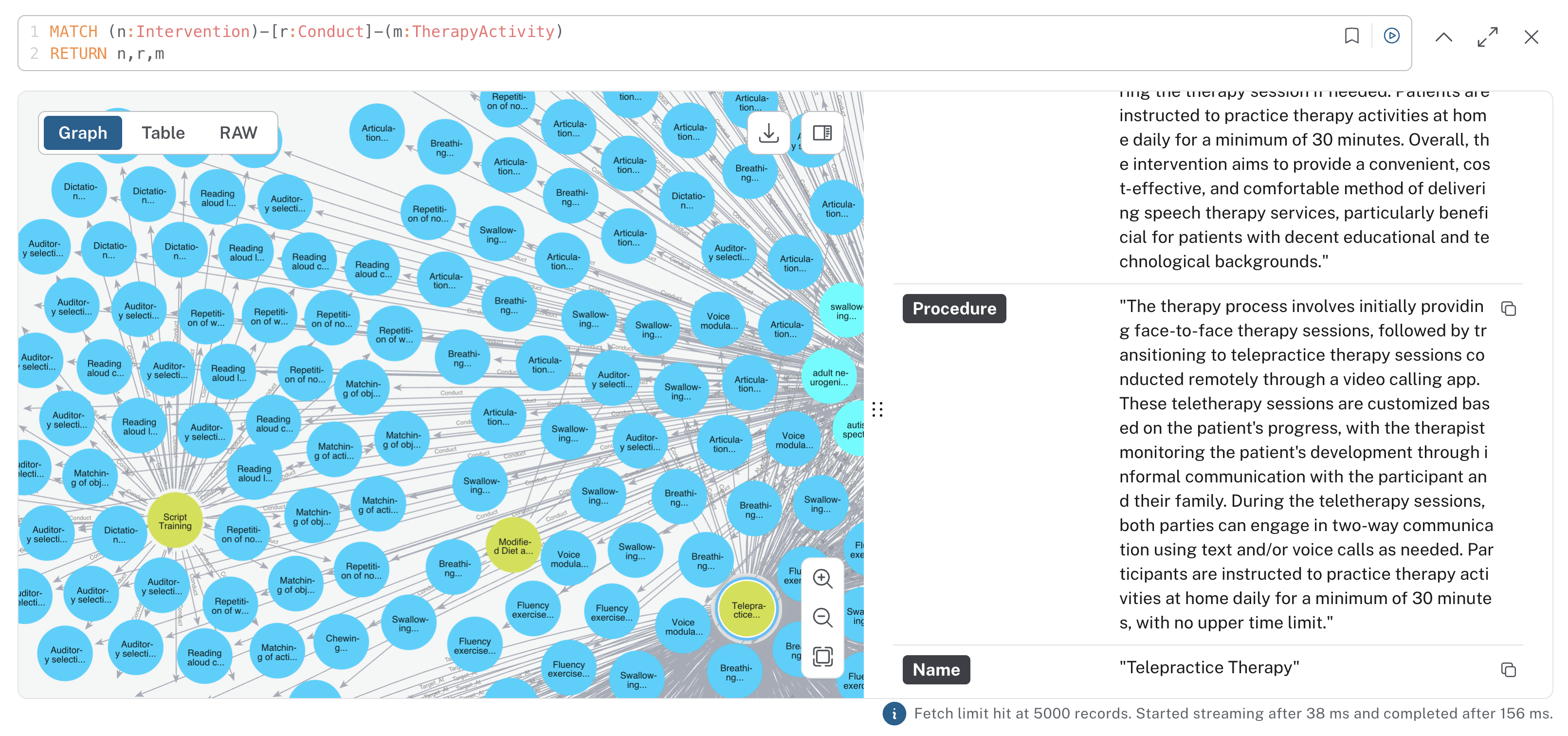}
    \caption{Partial views of Intervention-TherapyActivity in intervention KG.}
    \label{fig:kg2}
\end{figure*}

Through our intervention discovery framework, we constructed the first intervention knowledge graph in the speech-language pathology domain. Our intervention KG contains 33,148 nodes with 324,707 relations. We present some views of our intervention KG in Figure~\ref{fig:kg1} and Figure~\ref{fig:kg2}.
\end{document}